\documentclass{article}

\usepackage{arxiv}

\usepackage[utf8]{inputenc} 
\usepackage[T2A,T1]{fontenc}
\usepackage{hyperref}       
\usepackage{url}            
\usepackage{booktabs}       
\usepackage{amsfonts}       
\usepackage{nicefrac}       
\usepackage{microtype}      
\usepackage{lipsum}
\usepackage{amsmath}
\usepackage{amssymb}
\usepackage{xcolor}
\usepackage[super]{nth}
\usepackage[inline]{enumitem}
\usepackage{graphicx}
\graphicspath{ {./images/} }
\DeclareMathOperator*{\argmin}{arg\,min}

\title{Type Prediction Systems}

\author{
 Sarthak Dash \\
  IBM Research AI\\
  Yorktown Heights, NY\\
  \texttt{sdash@us.ibm.com} \\
   \And
  Nandana Mihindukulasooriya \\
  IBM Research AI\\
  Yorktown Heights, NY\\
  \texttt{nandana.m@ibm.com} \\
  \And
 Alfio Gliozzo \\
  IBM Research AI\\
  Yorktown Heights, NY\\
  \texttt{gliozzo@us.ibm.com} \\
  \And 
  Mustafa Canim \\ 
  IBM Research AI\\ 
  Yorktown Heights, NY\\
  \texttt{mustafa@us.ibm.com}
}

\begin{document}
\maketitle
\begin{abstract}
Inferring semantic types for entity mentions within text documents is an important asset for many downstream NLP tasks, such as Semantic Role Labelling, Entity Disambiguation, Knowledge Base Question Answering, etc. Prior works have mostly focused on supervised solutions that generally operate on relatively small-to-medium-sized type systems. In this work, we describe two systems aimed at predicting type information for the following two tasks, namely \begin {enumerate*} [label=\itshape\alph*\upshape)] \item A \textsc{TypeSuggest} module, an \emph{unsupervised} system designed to predict types for a set of user-entered query terms, and \item An Answer Type prediction module, that provides a solution for the task of determining the correct type of the answer expected to a given query. \end{enumerate*} Our systems generalize to arbitrary type systems of \emph{any} sizes, thereby making it a highly appealing solution to extract type information at any granularity.
\end{abstract}

\section{Introduction} 

Knowledge graphs are generally defined as ``\textit{a graph of data intended to accumulate and convey knowledge of the real world, whose nodes represent entities of
interest and whose edges represent relations between these entities}"~\cite{hogan2020knowledge}. Knowledge graphs are designed as directed edge-labelled graphs (also known as multi-relational graphs) where each edge has a direction and a relation type. Most of them use Semantic Web technologies such as RDF(S) and OWL for representing these knowledge graphs with explicit semantics. RDF type or ``\textit{is-a}" relation of relation is one very important relation type that defines the type of each node. For example, if the node is ``\textit{New York}", it could be typed as a ``\textit{city}" or ``\textit{populated city}".

Identifying these types can provide vital information in many knowledge graph related tasks such as Knowledge Base Question Answering (KBQA)~\cite{hoffner2017survey}, Table Understanding~\cite{zhang2020web}, Table Question Answering (Table QA), etc. For e.g.\ when a question is asked by the user corresponding to a KBQA or a Table QA task, it is crucial to understand the type of the correct answer pertaining to the question before proceeding to answer search. 

In this paper, we introduce two systems for predicting type information, namely \begin {enumerate*} [label=\itshape\alph*\upshape)] \item The \textsc{TypeSuggest} module, an \emph{unsupervised} system designed to predict types for a set of user entered seed query terms, and \item The Answer Type prediction module, that \emph{pulls-in} types from the Semantic Web in a distantly supervised fashion, to predict the type of the correct answer for the user provided question. \end{enumerate*} The input to the \textsc{TypeSuggest} module is a list of seed query terms, and the output returned by the module is a ranked list of types from some predefined type system. The Answer Type Prediction module takes in question string as input, and returns a ranked list of types for the correct answer. Both of these systems \emph{do not} require manual annotations, thereby making them highly appealing to be used as it is, in a wide-variety of domains.
\section{Related Work} 
\label{sec:related work}

There have been many studies over the years focusing on this problem. In this work, we limit the discussion to those that go beyond predicting a fixed set of \emph{coarse-grained} answer types.

One of the early works \cite{li2002learning} involved using fine-grained and coarse-grained types together in a hierarchical classifier, albeit the types were \emph{not} overlapping. Other works such as \cite{khoury2011question, lopez2012poweraqua} use hand-crafted features or features obtained from pre-existing grammars in conjunction with machine learning models (such as SVMs, Logistic Regression, etc.) to build answer type prediction models. Furthermore, \cite{choi2018ultra} uses additional manual annotations to perform fine-grained typing of entity mentions within a sentence. More recently, SeMantic AnsweR Type prediction task (SMART)~\cite{smart2020} was proposed as an ISWC 2020 Semantic Web challenge in which eight answer type systems participated. The top system at SMART 2020 formulates the type prediction task as an extreme multi-label text classification (XMC) problem and uses X-Transformers~\cite{chang2020taming} for type prediction. 

In comparison, our proposed method utilizes \emph{types} from the Semantic Web in a distantly-supervised fashion, thereby requiring no manual annotations. In our approach, one can also determine the granularity level of the final \emph{types}, since these are obtained from open domain ontologies.

\section{Type Suggestion Module} \label{sec:type_predict}

%

In this section, we introduce the \textsc{TypeSuggest} module designed to generate types for a set of seed query terms input by the user. It is also a \emph{pre-requisite} needed to build the \textsc{Answer Type Prediction} module. In essence, given a set of seed terms as input, \textsc{TypeSuggest} provides a ranked list of \emph{relevant types} as output. Implementation wise, it uses a predefined type system $\mathcal{TS}$ such as DBPedia, Wikidata, or Yago as a source of potential \emph{possible} types.

Given a list of seed query terms, $\mathcal{Q} = \{\emph{Amazon, Sony, IBM}\}$, in order to figure out relevant \emph{types}, \textsc{TypeSuggest} employs the following steps:

\paragraph{\textbf{Entity Linking:}}
The first step links the terms in $\mathcal{Q}$ to a taxonomy in $\mathcal{TS}$. This is done by first normalizing the string and then checking similarity with the entity labels in $\mathcal{TS}$. Once complete, we get a list $\mathcal{L}_{\mathcal{S}}$ of seed terms that are linked to their corresponding entity within $\mathcal{TS}$. The query terms that can not be linked are ignored.

\paragraph{\textbf{Seed Expansion:}} The second step is to expand the seed terms $\mathcal{L}_{\mathcal{S}}$, if necessary. The minimum number of seed terms required to facilitate next step is denoted by $k$ (typically set by user). Given the seed terms $\mathcal{L}_{\mathcal{S}}$, an expansion step (performed only if $|\mathcal{L}_\mathcal{S}| < k$) uses a pretrained \texttt{Word2Vec} \cite{mikolov2013distributed} model (with vocab $\mathcal{V}$) to augment $\mathcal{L}_\mathcal{S}$ in the following way: For each $x \in \mathcal{L}_\mathcal{S}$, it identifies the most similar term $y \in \mathcal{V} \setminus \mathcal{L}_\mathcal{S}$ which links to a valid entity in $\mathcal{TS}$. This term $y$ gets added to $\mathcal{L}_\mathcal{S}$, and the process is repeated until $|\mathcal{L}_\mathcal{S}| = k$.

\paragraph{\textbf{Type Identification:}} The final step is to identify the types based on the linked seed terms. For each linked seed term, we look for its \emph{type} (denoted by the function \textsc{Type}) within the taxonomy $\mathcal{TS}$, thus yielding a \emph{type-vocabulary} $\mathcal{TV}$. For each \emph{type} $t^\prime \in \mathcal{TV}$, we define,
\begin{equation*}
    \textsc{Count}(t^\prime) = |\{x \in \textsc{SeedExpansion}(\mathcal{L}_\mathcal{S}) | \textsc{Type}(x)=t^\prime\}|
\end{equation*}

Furthermore, for each $t^\prime \in \mathcal{TV}$, we define \textsc{EntityCount}($t^\prime)$ as the number of entities associated with the \emph{type} $t^\prime$ in $\mathcal{TS}$. Moreover, let $Z(t^\prime)$ denote the total number of entities of type $t^\prime$ present in $\mathcal{TS}$. 

We rank the \emph{types} in $\mathcal{TV}$ using the following tf-idf like function,
\begin{equation*}
    \forall t^\prime \in \mathcal{TV}, \quad \textsc{Score}(t^\prime) = \textsc{Count}(t^\prime) \times [\log Z(t^\prime) - \log \textsc{EntityCount}(t^\prime)]
\end{equation*}
and return the ranked list as the output of the \textsc{TypeSuggest} module.

\section{Answer Type Prediction}\label{sec:answerType}
In this section, we use the \textsc{TypeSuggest} module described in the previous section to build our proposed Answer Type Prediction model. 

As input, our model takes a list of question-answer pairs, i.e.\ $\zeta=(q_i, a_i), 1\leq i \leq N$ from a large external general-purpose QA dataset. This list of question-answer pairs $\zeta$ goes through the following pre-processing steps.

\paragraph{\textbf{Type Acquisition:}}
For each question-answer pair $(q_i, a_i)$, we define the \emph{type} of the answer $a_i$ as follows,
\begin{equation} \label{type_def}
    \textsc{Type}(a_i) = 
    \left\{
        \begin{array}{ll}
             [(\textsc{NER}(a_i), \textsc{Score})] & \quad \mbox{if } \textsc{NER}(a_i) \in \mathcal{N}  \\
             \textsc{TypeSuggest}(a_i) & \quad \mbox{elsewhere} 
        \end{array}
    \right.
\end{equation}

where $\mathcal{N} = \texttt{\{Date, Cardinal, Ordinal, Quantity, Money, Percent\}}$ is a set of coarse named entity types, NER denotes an off-the-shelf Named Entity Recognition System and \textsc{Score} is a fixed numerical value, i.e.\ a \emph{hyper-parameter} of our proposed model.

Using this criterion, we obtain a ranked list of \emph{types} $T_i$ for each question-answer pair $(q_i, a_i)$, and construct our augmented dataset, $\zeta^{\textsc{Aug}} = (q_i, a_i, T_i)$. Note that $\forall i, 1\leq i \leq N, T_i$ denotes a ranked list of \emph{types}.

\paragraph{\textbf{Type Restriction:}}
The augmented dataset $\zeta^{\textsc{Aug}}$ built in the previous step is highly likely to yield a large number of unique \emph{types} because the predefined type systems $\mathcal{TS}$ are usually very large in size. In this step, we restrict ourselves to top $k$ (by frequency) unique types and drop the rest. Here, $k$ is a hyper-parameter of our model.

Out of the \emph{types} that remain, we keep only those  whose frequency is less than a pre-set threshold count $c$, thus ensuring that the \emph{generic} types such as \texttt{yago:Thing}, \texttt{yago:Abstraction} etc.\ are removed, which allows the \emph{learner} (to be described later) not to be \emph{biased} towards these generic types. 

Let $\mathcal{T}$ denote the union of the left-over \emph{types} with the set $\mathcal{N}$ defined above. This set constitutes our \emph{Type Vocabulary},i.e.\ we build a model (described below) which given a question $q$ as input, returns a ranked list of entries from $\mathcal{T}$ as output. 

This \emph{frequency} based pruning step transforms the augmented dataset $\zeta^{\textsc{Aug}}$ in the following fashion. For each entry $(q_i, a_i, T_i) \in \zeta^{\textsc{Aug}}$, wherein $T_i= (t^j_i, s^j_i),\text{} 1\leq j \leq |T_i|$ denotes the ranked list of types corresponding to answer $a_i$, we have the following update,

\begin{equation}\label{type_restrict}
    \tilde{t}^j_i = 
    \left\{
        \begin{array}{ll}
             t^j_i & \quad \mbox{if } t^j_i \in \mathcal{T}  \\
             \argmin_{x^{*} \in \mathcal{T}} \textsc{Dist}(x, x^{*}) & \quad \mbox{if } t^j_i \notin \mathcal{T}
        \end{array}
    \right.
\end{equation}
wherein \textsc{Dist} between two types $x$ and $x^{*}$ is defined as the length of the shortest path between types $x$ and $x^{*}$ within the type system $\mathcal{TS}$ encoded as a graph.

Thus, this step outputs a pre-processed dataset $\zeta^{\textsc{Aug}}_{\textsc{Restr}} = (q_i, a_i, \tilde{T}_i)$, wherein each instance now contains an updated list of \emph{types} $\tilde{T}_i = (\tilde{t}^j_i, s^j_i),\text{} 1\leq j \leq |\tilde{T}_i|$ for answer string $a_i$ according to Equation \ref{type_restrict}.

This pre-processed dataset $\zeta^{\textsc{Aug}}_{\textsc{Restr}}$ forms the training dataset for our proposed \emph{neural network} architecture for the \emph{Answer Type Prediction} task, which is described in detail in the next section. 

\paragraph{\textbf{Neural Network Architecture:}}

\begin{figure}[ht]
\centering 
\includegraphics[width=0.8\linewidth]{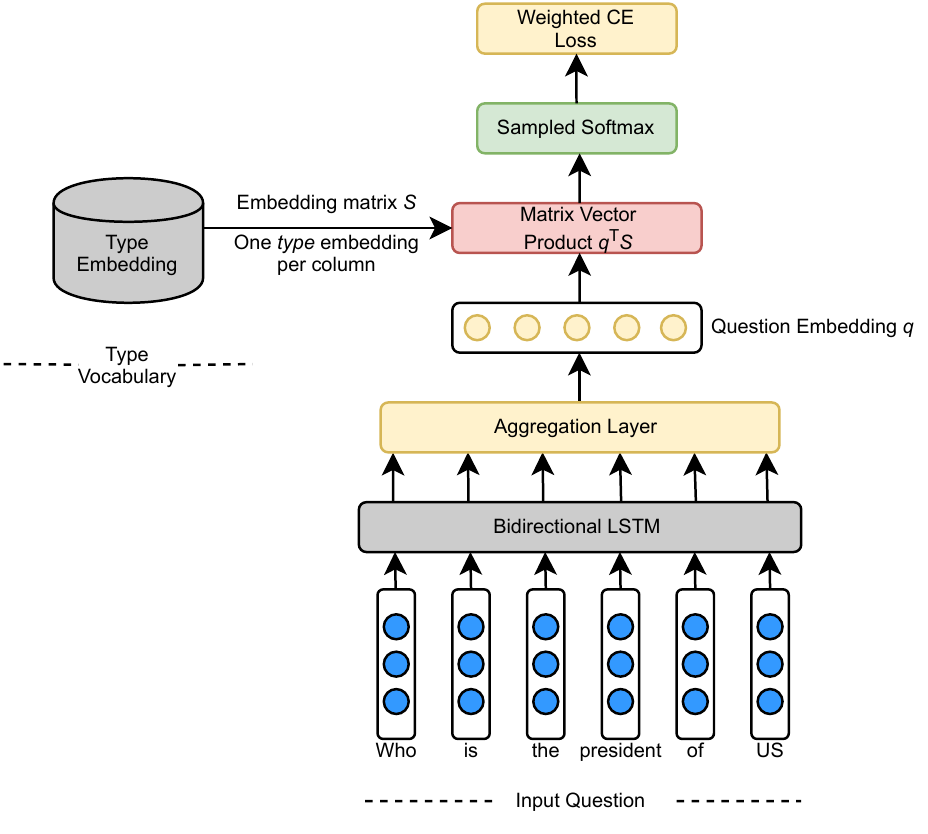}
\caption{Architecture of \emph{Answer Type Prediction} model}
\label{fig.qtype_classifier}
\end{figure}

We next describe the detailed architecture of our proposed \emph{Answer Type Prediction} model. As shown in Figure~\ref{fig.qtype_classifier}, it consists of three steps: 

\begin{itemize}
    \item Preparing \emph{Type Embeddings} from the Type vocabulary $\mathcal{T}$.
    \item Encoding input \emph{question} $q_i$ to its corresponding question embedding $\vec{q}_i$.
    \item Building a simple learning framework that uses $\vec{q}_i$ and $\mathcal{T}$ as inputs, and produces a list of ranked \emph{types} $T^{\textsc{Pred}}_i$ as output. 
\end{itemize}

The predicted list of \emph{types} $T^{\textsc{Pred}}_i$ is compared with $\tilde{T}_i$ (from $\zeta^{\textsc{Aug}}_{\textsc{Restr}}$) in order to calculate \emph{Weighted Cross Entropy} loss values, which is minimized during training procedure. We now describe each of these \emph{three} sections in detail.

\paragraph{\textbf{Preparing Type Embeddings:}} We use a pre-trained \textsc{Glove} model \cite{pennington2014glove}, and encode each entry within the Type vocabulary $\mathcal{T}$ to initialize a Type Embedding matrix $S$. Let us illustrate via an example.
\begin{itemize}
    \item If the type within $\mathcal{T}$ is \texttt{dbo:person} or \texttt{Wikidata:person}, we assign it the \textsc{Glove} vector for ``person''.
    \item If the type is \texttt{yago:WikicatAmericanPeople}, this type gets assigned the vector $0.5 * [v(\texttt{american})+v(\texttt{people})]$, where $v(\texttt{american})$ and $v(\texttt{people})$ denote the \textsc{Glove} embeddings for the terms ``american'' and ``people'' respectively. 
    
    We split \texttt{WikicatAmericanPeople} to \texttt{Wikicat}, \texttt{American}, and \texttt{People} using regular expressions. The term \texttt{Wikicat} is ignored.

    \item If the type is \texttt{yago:}\texttt{Person100007846}, this type gets assigned the \textsc{Glove} vector for ``Person''. The numerical part (extracted via \emph{regex}) is ignored.
    \item Any \emph{term} not present in \textsc{Glove} gets initialized randomly.
\end{itemize}


\paragraph{\textbf{Preparing Question Embedding:}} 


As illustrated in Figure~\ref{fig.qtype_classifier}, given a training instance $(q_i, a_i, \tilde{T}_i) \in \zeta^{\textsc{Aug}}_{\textsc{Restr}}$, the question string is tokenized and encoded via a \emph{bi-directional} LSTM~\cite{hochreiter1997long}. The vectors from both the unidirectional LSTMs are concatenated together to yield a matrix $Q_i$ of dimensions $|q_i| \times 2D$, where $|q_i|$ denotes the number of tokens in the question string $q_i$ and $D$ is the size of the vectors output by the unidirectional LSTMs. 

The \emph{Aggregation} layer then consists of the following steps,
\begin{itemize}
    \item Calculate $\psi_i = Q_i \cdot \textsc{ATT}$ where matrix \textsc{ATT} is of dimensions $2D \times 1$, and $\cdot$ represents matrix multiplication.
    \item Normalize $\psi_i$ to have unit $l_2$ norm, and calculate $Q_i \otimes \psi_i$, where $\otimes$ denotes element-wise multiplication.
    \item Calculate $\vec{q}_i = \sum_{j=1}^{|q_i|} (Q_i \otimes \psi_i)_{[j,:]}$, i.e.\ sum together all the rows in the matrix obtained in the previous step.
    \item Here, \textsc{ATT} denotes the model \emph{parameters} learned during training.
\end{itemize}

Using these steps, the question string $q_i$ is mapped to its \emph{vector} representation $\vec{q}_i$. Next, we describe how $\vec{q}_i$ interacts with the Type Embedding matrix $S$.

\paragraph{\textbf{Putting it Together:}}
In this step, our model performs a \emph{matrix-vector} multiplication $\vec{q}^T_iS$, followed by a sampled softmax operation, in order to obtain a predicted list $\mathcal{P}_i$ of \emph{scores}, one per type. Using this predicted list $\mathcal{P}_i$ and the known list of types $\tilde{T}_i$, the loss function (also referred to as \emph{Weighted Cross Entropy} loss) is calculated as,

\begin{equation*}
    \mathcal{J}_i = - \sum_{(\tilde{t}^j_i, s^j_i)\in \tilde{T}_i} s^j_i \log \mathcal{P}^{\tilde{t}^j_i}_i
\end{equation*}

where $\mathcal{P}^{\tilde{t}^j_i}_i$ corresponds to the \emph{system predicted} score for the type $\tilde{t}^j_i$.

During inference time, given a query question $q_{\textsc{Query}}$, we use the learned model parameters to build the query embedding $\vec{q}_{\textsc{Query}}$ as well as the embedding matrix $S$. Finally, we perform a matrix-vector multiplication, i.e. $\vec{q}^T_{\textsc{Query}}S$ to obtain a ranked list of \emph{predicted} scores, which is shown to the user.

\section{Conclusion} 

In this paper, we introduced two systems for predicting type information, namely \begin {enumerate*} [label=\itshape\alph*\upshape)] \item The \textsc{TypeSuggest} module, an \emph{unsupervised} system designed to predict types for a set of user entered seed query terms, and \item The Answer Type prediction module, that \emph{pulls-in} types from the Semantic Web in a distantly supervised fashion, to predict the type of the correct answer for the user provided question. \end{enumerate*} We described both of these systems in detail, starting with the data ingestion phase, followed by the pre-processing phase and ended up with a neural network based learner for the Answer Type prediction module. Furthermore, we demonstrated that both these systems \emph{do not} require manual annotations, thereby making them highly appealing to be used as it is, in a wide-variety of domains.

\bibliographystyle{acm}
\bibliography{bibliography}

\end{document}